\documentclass[10pt,twocolumn,letterpaper]{article}

\usepackage{cvpr}              
\usepackage{bbding}
\usepackage{pifont}
\usepackage{multirow}
\usepackage{siunitx}
\usepackage[dvipsnames]{xcolor}
\usepackage{amsthm}
\usepackage{bm}

\newcommand{\longlongname}{Mutual Information-aware Multimodal Iterated Relational dAta GEneration (MI$^{2}$RAGE)}
\newcommand{\shortname}{MI$^{2}$RAGE}
\newtheorem{proposition}{Proposition}

\definecolor{cvprblue}{rgb}{0.21,0.49,0.74}
\usepackage[pagebackref,breaklinks,colorlinks,citecolor=cvprblue]{hyperref}
\usepackage{soul}
\usepackage{amssymb,amsthm}
\usepackage{verbatim}
\def\paperID{17304} 
\def\confName{CVPR}
\def\confYear{2024}

\title{Training on Synthetic Data Beats Real Data in Multimodal Relation Extraction}

\author{Zilin Du, Haoxin Li, Xu Guo, Boyang Li\\
Nanyang Technological University\\
Singapore\\
{\tt\small \{zilin003, haoxin003, xu008, boyang.li\}@ntu.edu.sg}
}

\begin{document}
\maketitle
\begin{abstract}

The task of multimodal relation extraction has attracted significant research attention, but progress is constrained by the scarcity of available training data. One natural thought is to extend existing datasets with cross-modal generative models. In this paper, we consider a novel problem setting,  where only unimodal data, either text or image, are available during training. We aim to train a multimodal classifier from synthetic data that perform well on real multimodal test data. However, training with synthetic data suffers from two obstacles: lack of data diversity and label information loss. To alleviate the issues, we propose \longlongname, which applies Chained Cross-modal Generation (CCG) to promote diversity in the generated data and exploits a teacher network to select valuable training samples with high mutual information with the ground-truth labels. Comparing our method to direct training on synthetic data, we observed a significant improvement of 24.06\% F1 with synthetic text and 26.42\% F1 with synthetic images. Notably, our best model trained on completely synthetic images outperforms prior state-of-the-art models trained on real multimodal data by a margin of 3.76\% in F1. Our codebase will be made available upon acceptance.
\end{abstract}    
\section{Introduction}
\label{sec:intro}


Relation extraction (RE) \cite{xiaoyan2023comprehensive,yan2022empirical,zeng2020copymtl,li2019entity,fu2019graphrel} aims to categorize relationships between two entities. 
Recently, there has been significant interest in multimodal relation extraction (MRE) \cite{zheng2021mnre, chen2023unified, hu2023multimodal, chen2023chain, cui2023enhancing, li2023dual, zhao2023tsvfn}. Compared to text-only relation extraction, MRE offers the advantages of reduced ambiguity and enhanced representation learning by utilizing complementary bimodal features. However, advances in MRE are hindered by data scarcity, partially due to the difficulty of collecting well-aligned multimodal data. For example, the popular MNRE-2 dataset \cite{zheng2021mnre} contains 
15,485 data samples in 
only 23 relation types. In contrast, several textual RE datasets, such as WebNLG \cite{gardent2017creating},  WikiReading \cite{hewlett2016wikireading}, and FewRel \cite{gao2019fewrel}, possess more than 100 relation classes and up to millions of data instances.


To handle data scarcity in MRE, a natural thought is to complement unimodal training data with synthetic data to create multimodal training data, since text-to-image generative models \cite{ho2020denoising, rombach2022high, saharia2022photorealistic, ramesh2022hierarchical, balaji2022ediffi, kumari2023multi, zhang2023adding,dai2023emu} and image-to-text captioning models \cite{OSCAR-2020, zhang2021vinvl, wang2021simvlm, alayrac2022flamingo, li2023blip, wang2022ofa, Singh_2022_CVPR, liu2023improvedllava} have made impressive stride recently. To investigate the feasibility of this approach, we consider a new problem, Multimodal Relation Extraction with a Missing Modality, where we artificially remove one modality from the training data. We train a multimodal network using real data in one modality and synthetic data in the other. During inference, we use real data from both modalities, if they are available. 

\begin{figure*}[t]
  \centering
  \includegraphics[width=1.0\linewidth]{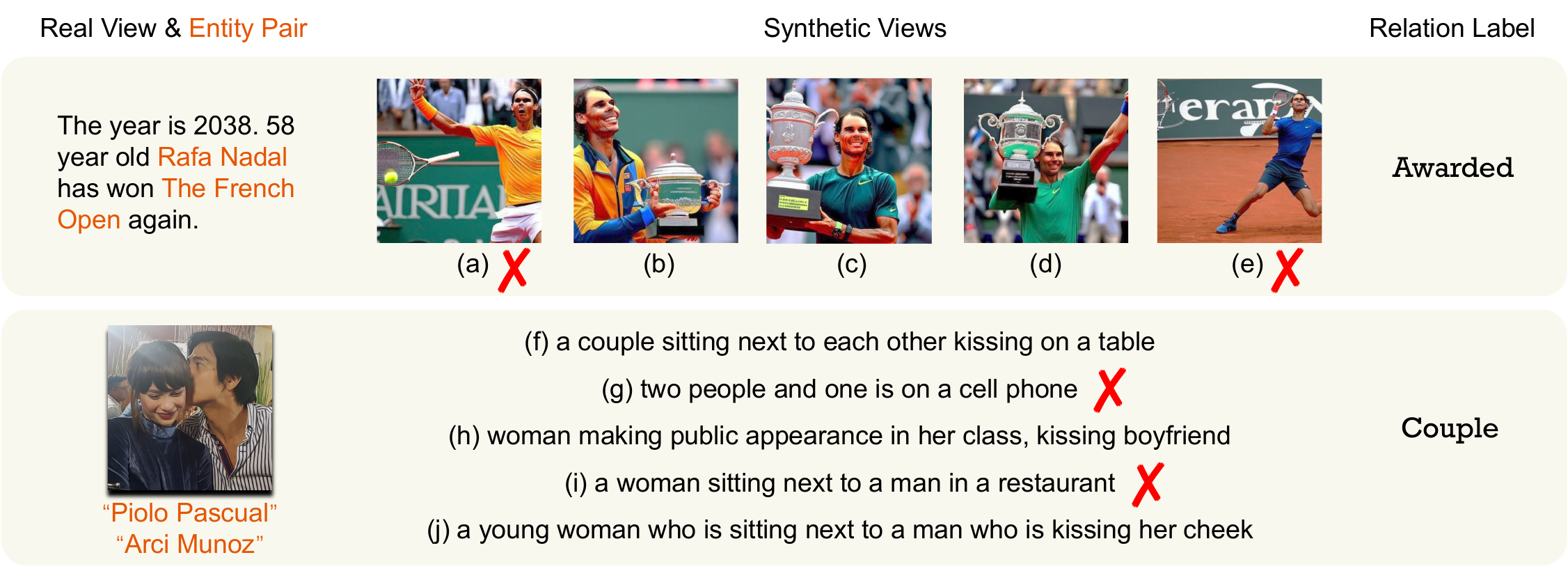}
  \caption{Examples of synthetic data from cross-modality generative models. The first row is image generation, while the second row is image captioning. The samples marked by {\color{red}\ding{56}} are unrelated to the relation label.}
  \label{IMGExample}
\end{figure*}

However, training neural networks with synthetic data is fraught with challenges. In this paper, we identify and tackle two, namely the lack of data diversity and label information loss. The first challenge, the lack of diversity, is caused by the fact that generative models tend to over-represent high-frequency content \cite{Freq_bias_GAN_2021, Hendricks_2018_ECCV}, leading to suppressed tails of the data distribution. This is also observable in Fig. \ref{IMGExample}, where the generated images appear rather similar. To alleviate this issue, we propose Chained Cross-modal Generation, where we chain and repeatedly apply text-to-image and image-to-text generators. For example, we first  generate some text from synthetic images, and generate images on the resulting text. Each generation step introduces some variance to the generated data and enhances diversity. 

\begin{figure}[t]
  \centering
  \includegraphics[width=0.85\linewidth]{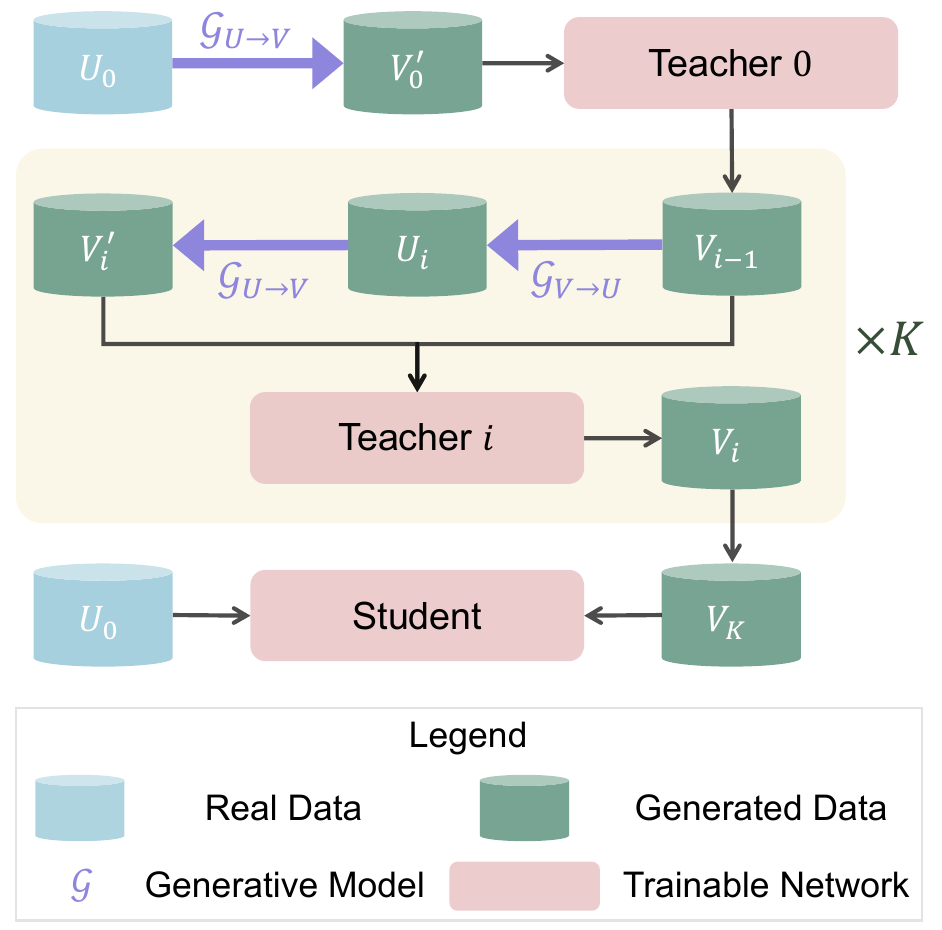}
  \caption{Overview of \shortname{}. $U$ denotes real data in one modality (text or imagery) and $V$ denotes synthetic data in the other modality. Starting from real data $U_0$, we iteratively apply generative networks $\mathcal{G}_{U\rightarrow V}$ and $\mathcal{G}_{V\rightarrow U}$ to generate a series of synthetic datasets $V_i$. We train teacher networks to select informative training data. The student network is trained with the real views $U_0$ and the selected synthetic views $V_k$. } 
  \label{IMGoverview}
\end{figure}

Another obstacle is that, synthetic data, especially after repeated Chained Cross-modal Generation, may lose information about the data label. We refer to this issue as label information loss. Some examples are given in Fig. \ref{IMGExample} (a) and (e), where the synthetic image becomes unrelated to the relation label \texttt{Awarded}. To handle the issue, we train a (mostly) unimodal teacher network to predict the label from the synthetic modality only. After that, we keep only data instances with low training losses, which are then used either as input to the next chained generation round or as training data for a multimodal student network (See Fig. \ref{IMGoverview}). This process can be understood as identifying synthetic data with high mutual information with ground-truth labels (see detailed discussion in \S \ref{sec:why-teacher}), which are intuitively good data for training \cite{tishby99information,hjelm2018learning}.

We propose \longlongname, which iteratively applies cross-modal generators to create a rich and diverse multimodal training set and exploits a teacher network to select valuable training samples with high mutual information with the ground-truth labels. Our contributions can be summarized as follows.
\begin{itemize} 
\item We investigate a novel and challenging variation of multimodal relation extraction where, during training, data from one modality are inaccessible  and must be synthesized using cross-modal generation. 
\item We propose to chain text-to-image and image-to-text generators in order to enhance data diversity and employ a teacher network to select training samples with low training losses. From an information-theoretical perspective, we justify the teacher network as selecting synthetic data that are informative about the label. 
\item Our best model, trained on real text and synthetic images, sets a new state of the art on MNRE-2 and outperforms the best baseline, TMR \cite{zheng-etal-2023-rethinking}, by 3.76\% in F1, even though TMR was trained on both real and synthetic data in both modalities. An ablation study reveals that the proposed technique surpasses naive synthetic data baselines by a massive 24.28-26.52\% margin in F1 score. 
 \end{itemize}


\section{Problem Definition}
\label{sec:task_definition}

Multimodal relation extraction begins with a text snippet, which contains textual references to two entities, and an image. Between the entities, it is necessary to differentiate between the \emph{subject} and the \emph{object}, as the relation between them may be unidirectional (\emph{e.g.}, \texttt{Above}, \texttt{Eat}). An example subject-relation-object tuple is (\texttt{Yoda}, \texttt{Present-in}, \texttt{Last Jedi}). Given the input, the task is to predict the exact relation between the two entities, but they may also be unrelated. 

In this paper, we propose a new problem formulation, Multimodal Relation Extraction with a Missing Modality (MREMM), in which data from one modality are inaccessible during training. Specifically, we investigate two settings: \textit{Without Real Texts} and \textit{Without Real Images}. Nevertheless, we need to learn a multimodal relation classifier. For simplicity, we often refer to the text snippet and the corresponding image as the two \emph{views} within the same data instance. One possible solution to MREMM is to synthesize the missing view from the available view, and use both as training data.  




\section{Method}
\label{sec:method}

Fig. \ref{IMGoverview} illustrates the overall procedure of \shortname. In the MREMM problem, we start with real training data from either the text or image modality, which is referred to as $U_0$. After we feed $U_0$ to the cross-modal generator, we acquire a set of synthetic views called $V_0^\prime$. Note that we may generate several synthetic views for each real view. The cross-modal generator from $U_0$ to $V_0^\prime$ is denoted as $\mathcal{G}_{U\rightarrow V}$ and the generator in the reverse direction is denoted as $\mathcal{G}_{V\rightarrow U}$.

To select informative training data, we construct a teacher network that learns to predict the ground-truth label from the synthetic view \emph{only}. Note that, in $V_0^\prime$, each synthetic view is created from one real view and is thus associated with a ground-truth label. When the synthetic data are images, we also feed the entity names to the network, so that the network knows which entity pair the relation is for. From all synthetic data, we select those with low training losses, yielding a new dataset $V_0$. We argue that the teacher network helps to select training data with high mutual information with the ground-truth label and guard against semantic drift. See analysis in \S \ref{sec:why-teacher}. 

Instead of stopping at $V_0$, we apply another round of cross-modal generation; we apply $\mathcal{G}_{V\rightarrow U}$ on $V_0$ and create $U_1$, which is then fed to $\mathcal{G}_{U\rightarrow V}$ to produce $V_1^{\prime}$. We then train another teacher network on $V_1^{\prime} \cup V_0$ to once again remove data points with high training loss. This process (yellow box in Fig. \ref{IMGoverview}) can be repeated $K$ times. We name this recursive generation Chained Cross-modal Generation. Finally, we combine the synthetic views $V_K$ and the real views $U_0$ as training data for the student network.

\subsection{Why the Teacher Networks}
\label{sec:why-teacher}
We justify the teacher networks from an information-theoretic perspective. With abuse of notation, we denote the real view (either text or image) as random variable $U$, and the synthetic view as random variable $V$. The class label is denoted as $Y$. The real-world data distributions are $P(\cdot)$ and the synthetic data distributions are $Q(\cdot)$. 
Note the off-the-shelf pretrained generator $\mathcal{G}_{U\rightarrow V}$ takes only $U$ as input and not the class label $Y$. Therefore, the synthetic data distribution factorizes as $Q(V|U)P(U|Y)$. More generally, as in the proposed Chained Cross-modal Generation, we chain $\mathcal{G}_{U\rightarrow V}$ and $\mathcal{G}_{V\rightarrow U}$ and generate a sequence of synthetic data, $V_1, U_2, V_2, \ldots, U_K, V_K$, whose distribution factorizes as
\begin{equation}
\label{eq:factorization}
P(U_1|Y)Q(V_1|U_1) \prod_{i=2}^K Q(U_i|V_{i-1})Q(V_i|U_i).
\end{equation}
However, each generation step may lose information of the label $Y$, so the final $V_K$ may be close to being independent of $Y$ and cannot serve as effective training data. Similar phenomena are sometimes referred to as semantic drift in the NLP literature \cite{zhang-bansal-2019-semantic-drift}. 

To prevent information loss, we aim to choose synthetic training data with high mutual information with the ground-truth labels. It is well known that minimizing the cross-entropy loss can be understood as maximizing mutual information $I(V;Y)$ between network input $V$ and label $Y$ 
(\emph{e.g.}, \cite{pmlr-v97-poole19a,Oord2018RepresentationLW,Tschannen2019OnMI,McAllester2020,qin2020rethinking,Malik-MI-2020}).
\begin{proposition}
\label{prop:mi-xe-loss}
Let $V$ be the input variable, $Y$ be a uniformly distributed discrete label, and $f(v, y) \in \mathbb{R}$ be an arbitrary neural network. The negative cross-entropy loss is a lower bound for the mutual information,
\begin{equation} \label{eq:mi-xe-loss}
I(V, Y) \ge \mathbb{E}_{V, Y}\left[\log \frac{\exp f(v, y)}{\sum_{y^{\prime}}[\exp f(v, y^{\prime})]}\right].
\end{equation}
\end{proposition}
\noindent Therefore, to find data with high mutual information with the label, we pick those with low training losses. 
Further, due to the Markov factorization of Eq. \ref{eq:factorization}, we can prove the following (details in the supplemental material).
\begin{proposition}
\label{prop:mi}
The mutual information $I(V_K;Y)$ is the lower bound of the mutual information between any pairs of variables on the Markov chain from $Y$ to $U_{K-1}$ :
\begin{equation} \label{eq:mi}
\begin{split}
I(V_K;Y) \le \min & \{I(U_i;Y), I(V_i;Y), \\ 
& I(V_i;U_i), I(U_i;V_{i-1})\} \quad \forall i. 
\end{split}
\end{equation}
\end{proposition}

\noindent Thus, if the teacher network finds $V_K$ with high mutual information with $Y$, we can ensure the information flows from $Y$ to $V_K$ through all the intermediate variables. On the other hand, if $V_i$ has low mutual information with $Y$, subsequent generation from $V_i$ cannot recover the information. Hence, we perform data filtering before every CCG iteration to make sure the synthesis of new data is conducted from sound starting positions. 

The use of separately trained teacher networks contrasts with some recent studies on synthetic training data \cite{rose2023visual, yang2022z, tang2023learning}, which discard text-image pairs with low similarity as determined by the pretrained CLIP model \cite{radford2021learning}. However, the two views may be similar in aspects unrelated to the label. For example, an image of a person wearing a red t-shirt holding a cup and the text ``a person wearing a red t-shirt'' may be quite similar, but the text omits the relation between the person and the cup and would have low mutual information with the label \texttt{Holding}. 

\subsection{The Teacher Network Architecture} 
\label{sec:method-transfer-teacher}

\begin{figure}[t]
  \centering
  \includegraphics[width=0.85\linewidth]{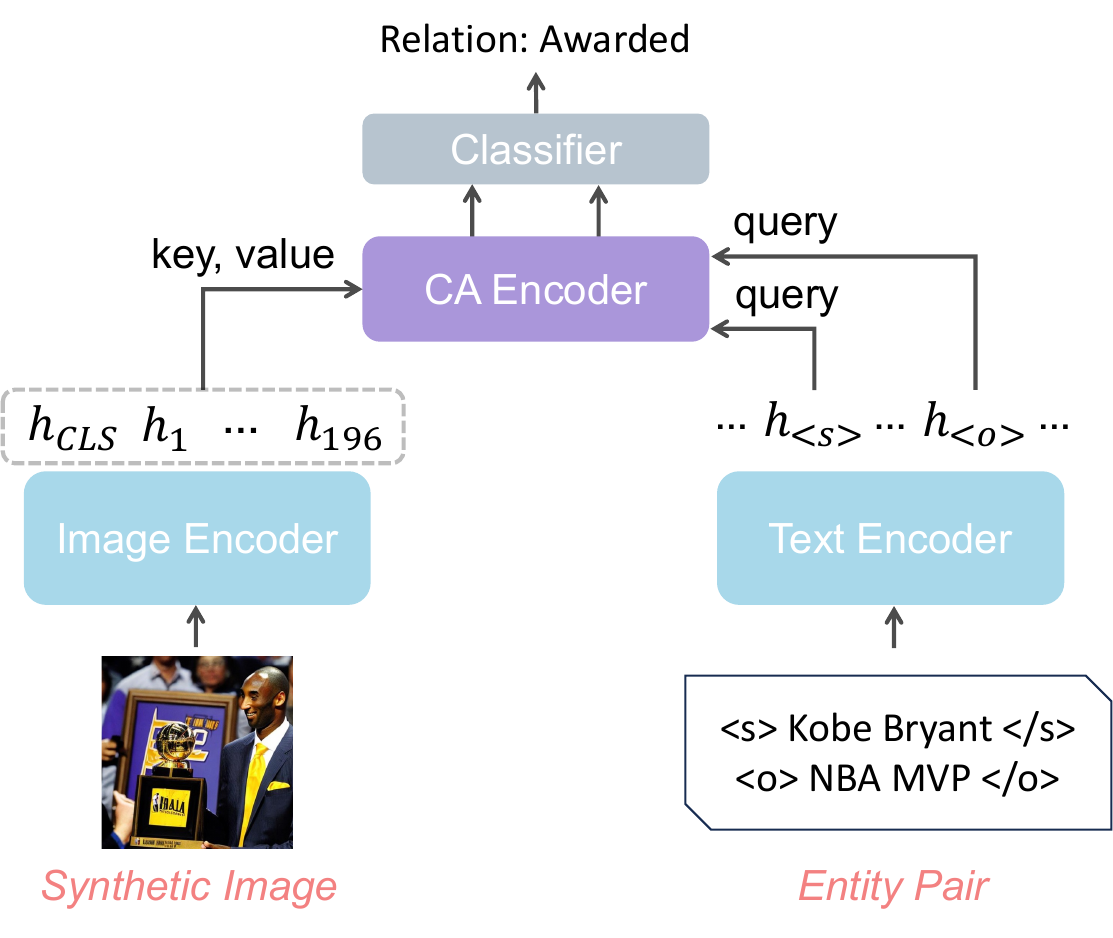}
  \caption{The teacher network for synthetic image selection. As input, the teacher receives only the entities, e.g., the subject \texttt{Kobe Bryant} and the object \texttt{NBA MVP}, and the synthetic image.} 
  \label{IMG_Teacher}
\end{figure}

\begin{figure*}[t]
  \centering
  \includegraphics[width=0.9\linewidth]{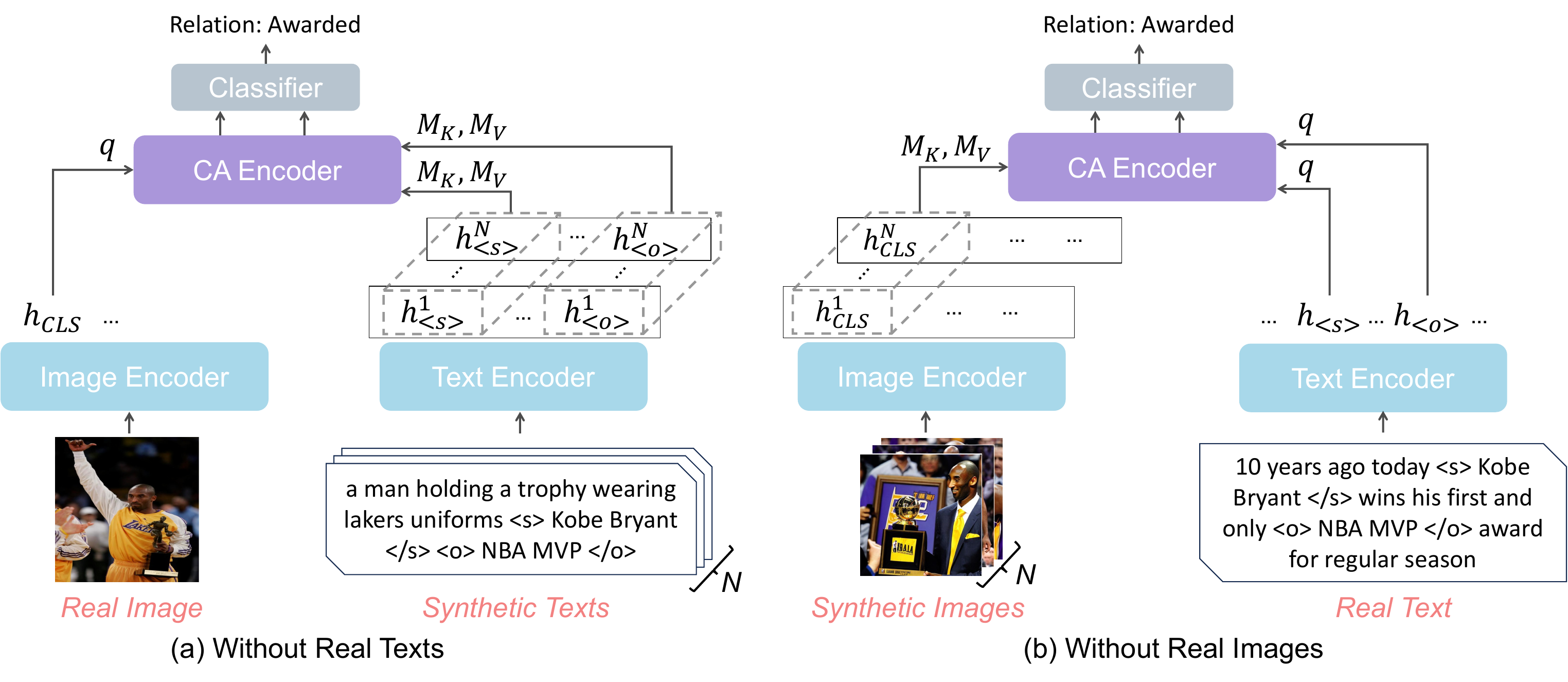}
  \caption{The student networks that predict the relation for an entity pair using real data from one modality and synthetic data from the other. (a) The student that is trained with a real image and $N$ synthetic texts. (b) The student that is trained with a real text snippet and $N$ synthetic images. } 
  \label{IMG_Student}
\end{figure*}

The objective of the teacher network is to select good synthetic data. As a result, we need to constrain its input to the synthetic view as much as possible. Suppose we provide both view to the teacher network, then it could learn to predict the label directly from the real view, which contains ample label information, in complete contradiction of the original purpose of the teacher network. 

Before introducing the teacher networks, we first discuss how the entities are encoded. Each entity pair contains a subject and an object. We use special tokens $\langle s \rangle$ and $\langle /s \rangle$ to mark the position of the subject entity and special tokens $\langle o \rangle$ and $\langle /o \rangle$ to mark the object. For example, the two-token subject [\texttt{New}, \texttt{York}] becomes [$\langle s \rangle$, \texttt{New}, \texttt{York}, $\langle /s \rangle$] after the special tokens are added. We then take the representations of the tokens $\langle s \rangle$ and $\langle o \rangle$ from the last encoder layer as the encodings of the subject and the object respectively, which are denoted as $\bm{h}_{\langle s \rangle}$ and $\bm{h}_{\langle o \rangle}$.

The teacher network responsible for filtering synthetic textual data is a pretrained text-only network, such as BERT. Its input contains a single image caption. We append both entity names at the end to remove any ambiguity since the synthetic captions often do not contain the entity names. A real text snippet, on the contrary, always contains the entity names. At the last layer, we extract the encoded entity vectors ($\bm{h}_{\langle s \rangle}$ and $\bm{h}_{\langle o \rangle}$ from $\langle s \rangle$ and $\langle o \rangle$), and feed their concatenation into a linear classifier. We finetune all neural network parameters using the cross-entropy loss. 

The teacher network for selecting synthetic images is slightly more complex; we show its architecture in Fig. \ref{IMG_Teacher}. We use a pretrained vision transformer to encode the image, and a pretrained textual encoder to incorporate information from the entities. We limit the input to the textual encoder to only the entity names, so that the network sees as little textual information as possible. We adopt a cross-attention layer, where the textual representations $\bm{h}_{\langle s \rangle}$ and $\bm{h}_{\langle o \rangle}$ serve as query vectors, and image patch encodings serve as the keys and values. The results of the cross-attention are two vectors, whose concatenation is input to a linear classifier. We finetune all network parameters.

\subsection{The Student Network Architecture} 
The student network should learn effective multimodal representations from real and synthetic training data. We design a network architecture that exploits a particular strength of training from synthetic data: we can generate as many synthetic views as we want for real view, both at training time and at inference time. Take text-to-image generation for example; each image may be considered as an imaginative interpretation of the text, which enriches the textual representation \cite{lu2022imagination, tan2020vokenization}. As such, we exploit the enrichment during both training and inference. 

The model architecture is shown in Fig. \ref{IMG_Student}. 
We use a cross-attention layer to combine multiple instances in the synthetic modality. Given a query vector $\bm{q}$, a key matrix $M_K$, and a value matrix $M_V$, the cross-attention operation can be written as 
\begin{equation}
\text{CA}(\bm{q}, M_K, M_V) = \sigma\left( \frac{M_K W_{K} W_{Q}\, \bm{q}}{\sqrt{d}}\right) (W_V M_V),
\end{equation}
where $\sigma(\cdot)$ is the softmax operation. $W_{K}, W_{Q}$ and $W_V$ are trainable parameters. $d$ is a scaling factor equal to the dimension  of $W_{Q}\, \bm{q}$. As in the original Transformer network, the results of cross-attention go through feedforward layers.   

When we process real image and synthetic texts (Fig.~\ref{IMG_Student}~(a)), for every image we receive $N$ captions from the teacher network and encode them separately. After that, we collect the subject encodings $\bm{h}_{\langle s \rangle}$ from all captions into $M^{sub}$, and all object encodings $\bm{h}_{\langle o \rangle}$ into $M^{obj}$. We take $\bm{h}_\text{CLS}$, the representation of the $\langle \text{CLS}\rangle$ token from the visual encoder, as the query vector. Applying the same CA operation twice, we have two vectors, 
\[
    \text{CA}(\bm{h}_\text{CLS}, M^{sub}, M^{sub}) \text{ and } \text{CA}(\bm{h}_\text{CLS}, M^{obj}, M^{obj}).
\]
The vectors go through the feedforward layers of the Transformer network. Their concatenation then go through a linear classifier. 

The case for synthetic images and real text is symmetrical (Fig.~\ref{IMG_Student}~(b)). For every text snippet, we acquire $N$ images from the teacher network, and encode them separately. After that, we take $\bm{h}_\text{CLS}$ of each image and arrange them into a matrix $M^{img}$. We take the entity representations $\bm{h}_{\langle s \rangle}$ and $\bm{h}_{\langle o \rangle}$ and use them as two query vectors. Again, we apply the CA operation twice and obtain two vectors,
\[
    \text{CA}(\bm{h}_{\langle s \rangle}, M^{img}, M^{img}) \text{ and }  \text{CA}(\bm{h}_{\langle o \rangle}, M^{img}, M^{img}),
\]
which are concatenated and input to subsequent Transformer layers and eventually to a linear classifier. The loss for both student networks is the cross-entropy loss. 

During inference, a test data instance contains two views and both are real. In order to fully realize the strength of data synthesis, we still generate a number of new views in the synthetic modality. Using the trained teacher network, we select $N$ synthetic views with the lowest losses. After that, we append the real view in the synthetic modality to the synthetic views, and feed all available views to the student network. This process is reminiscent of the technique of inference-time data augmentation. 

\subsection{Why Chained Cross-modal Generation}
\label{sec:ccg-justification}

To maximize the benefits from synthetic training data, they should be as similar to genuine data as possible. However, the synthetic data may exhibit significant differences from the real data. In particular, existing research \cite{Freq_bias_GAN_2021, Hendricks_2018_ECCV} suggests they may over-represent high-frequency content and have thinner tails than the real distribution. 
This phenomenon is also observable in the generated images in Fig. \ref{IMGExample}, which have similar backgrounds. 

We present a simple analysis of distribution similarity from the perspective of Monte Carlo expectation. The empirical negative cross-entropy can be understood as a Monte Carlo approximation for expectation,
\begin{equation}  
      \frac{1}{N} \sum_{y,v} \log P_{\theta}(Y=y|V=v) \approx \mathbb{E}_{P(V,Y)}[\log P_{\theta}(Y|V)],
\end{equation}
where $P_{\theta}(\cdot)$ denotes the prediction of a network parameterized by $\theta$, $N$ is the total number of data points, and $v$ and $y$ are drawn from the ground-truth distribution $P(V,Y)$. However, in practice $v$ and $y$ are drawn from the synthetic distribution $Q(V,Y)$. By importance sampling, we have
\begin{equation}
\label{eq:impt-sampling}
    \mathbb{E}[\log P_{\theta}(Y|V)] \approx \frac{1}{N} \sum_{y, v} \frac{P(Y|V)P(V)}{Q(Y|V)Q(V)} \log P_{\theta}(y|v).
\end{equation}
It is difficult to estimate $P(V, Y)$ and $Q(V, Y)$ and directly apply Eq. \ref{eq:impt-sampling}. Instead, we stick with the simple empirical cross-entropy loss, which works well only when $Q(Y|V)Q(V) \approx P(Y|V)P(V)$. In this paper, we encourage the similarity of $P(V)$ and $Q(V)$ using the CCG technique for sampling $V$. As each cross-modal generation step introduces some noise, chaining multiple generators likely results in diverse output.

\section{Experiments}
\label{sec:experiments}

\noindent\textbf{Dataset.} We use the MNRE-2 dataset \cite{zheng2021mnre}\footnote{https://github.com/thecharm/MNRE} in our experiments. It contains 15,484 samples within 9,201 image-caption pairs crawled from Twitter. There are 23 relation types including \texttt{None}. Following previous works, we split the dataset into training, validation and testing set with 12247, 1624 and 1614 samples, respectively. Meanwhile, we report the standard micro precision, recall and F1 score for evaluation. 

\vspace{0.5em}\noindent\textbf{Implementation Details.} The text-to-image generatior isStable Diffusion v2.1 \cite{rombach2022high} with 512×512 resolution, 50 denoising steps, and a guidance scale of 7.5. We use BLIP ViT-L \cite{li2022blip} as the image-to-text generator and set $p$ to 0.8 in nucleus (top-$p$) sampling \cite{holtzman2019curious}. 
For both the teacher and the student networks, the text encoder is a pretrained BERT-base and the vision encoder is CLIP ViT-B/16. The cross-attention module is a Transformer layer identical to BERT-base in architecture. 

The number of CCG iterations, $K$, is set to 2. For each real view in $U_0$, we first generate 30 synthetic views for $V_0^\prime$. The teacher network selects 60\%, which is equivalent to 18 views. The first CCG round generates 72 synthetic views, or 4 new views for each input. From the total of 90=18+72 views, another 60\% or 54 views are selected. The second CCG round generates another 54 views. Finally, we select 6 synthetic images for each real text and 10 captions for each real images for training the students. The number of selected synthetic views is determined on the validation set. At test time, we use the same generative procedure to produce synthetic views for each test instance. Further details are available in the supplementary.

\vspace{0.5em}
\noindent
\textbf{Baselines.} We compare our methods with 19 baselines, all of which are trained on the real multimodal data. \cite{zheng-etal-2023-rethinking, hu2023prompt} also leverage additional paired image-caption datasets, while \cite{wang2022named, hu2023multimodal} retrieve extra images and texts for training. For all baseline performance, we take the reported numbers from their respective paper directly. Details of the baselines can be found in the supplementary. 

\subsection{Main Results}
Table \ref{tab:main_results} compares \shortname{} with the baselines. Even though models trained on synthetic data are at an inherent disadvantage, our best model surpasses the best baseline, TMR \cite{zheng-etal-2023-rethinking}, by 3.76\% in F1 and the second best baseline, RECK \cite{feng2023towards}, by 3.97\% in F1 and 1.54\% in accuracy. Our model trained on synthetic text and real images also performs well, beating every baseline except for TMR and RECK in F1. In accuracy, this model snatches the third position, among all baselines that report this metric. 

We observe a performance gap between the synthetic image condition and the synthetic text condition. Based on our observation, we propose two hypothetical reasons. First, there is a distribution gap between synthetic text and real text. In a real text snippet, the entity names are part of the text and serve syntactic functions. However, the synthetic captions rarely contain the entity names, so we need to artificially append the entity names, creating a domain shift that may hurt generalization. Second, the images in MNRE-2 are sometimes  semantically misaligned with the text, which is well aligned with the label. Thus, synthetic text may deviate from the label and the real text more than synthetic images deviate from the label and the real images.


\begin{table}[!t]
\centering
  \caption{Main results on multimodal relation extraction.}
  \small
  \label{tab:main_results}
  \begin{tabular}{@{}ccccc@{}}
  
    \midrule
    Method & Accuracy & Precision & Recall & F1 \\
    
    \midrule
    VBERT \cite{li2019visualbert} & 73.97 & 57.15 & 59.48 & 58.30\\
    BSG \cite{zheng2021mnre} & 77.15 & 62.95 & 62.65 & 62.80\\
    UMT \cite{yu2020improving} & 77.84 & 62.93 & 63.88 & 63.46 \\
    UMGF \cite{zhang2021multi} & 79.27 & 64.38 & 66.23 & 65.29\\
    MEGA \cite{zheng2021multimodal} & 80.05 & 64.51 & 68.44 & 66.41\\
    MoRe \cite{wang2022named} & 79.87 & 65.25 & 67.32 & 66.27\\
    GPT4-XMLR \cite{chen2023chain} & - & - & - & 74.56\\
    DGF-PT \cite{li2023dual} & 79.82 & 79.72 & 78.63 & 79.24\\
    Iformer \cite{li2023analyzing} & 92.38 & 82.59 & 80.78 & 81.67\\
    HVPnet \cite{chen2022good} & 92.52 & 82.64 & 80.78 & 81.85\\
    TSVFN \cite{zhao2023tsvfn} & 92.67 & 85.16 & 82.07 & 83.02 \\
    MMIB \cite{cui2023enhancing} & - & 83.49 & 82.97 & 83.23\\
    MRE-ISE \cite{wu2023information} & 94.06 & 84.69 & 83.38 & 84.03\\
    I$^2$SRM \cite{huang2023i2srm} & - & 84.65 & 83.59 & 84.12\\
    VisualPT-MoE \cite{xu2023unified} & - & 84.81 & 83.75 & 84.28\\
    \cite{hu2023multimodal} & 93.54 & 85.03 & 84.25 & 84.64\\
    \cite{hu2023prompt} & - & 84.95 & 85.76 & 84.86\\
    RECK \cite{feng2023towards} & 95.11 & 88.77 & \underline{88.91} & 88.84\\
    TMR \cite{zheng-etal-2023-rethinking} & - & 90.48 & 87.66 & 89.05 \\
    \midrule
    \multicolumn{5}{c}{\textit{Without Real Texts}} \\
    \shortname{} ($K$ = 1) & 92.63 & 87.48 & 84.06 & 85.74  \\
    \shortname{} ($K$ = 2) & 93.60 & 87.92 & 84.96 & 86.42 \\
    \multicolumn{5}{c}{\textit{Without Real Images}} \\
    \shortname{} ($K$ = 1) & \underline{95.42} & \underline{93.74} & \underline{88.91} & \underline{91.26}  \\
    \shortname{} ($K$ = 2) & \textbf{96.65} & \textbf{93.92} & \textbf{91.72} & \textbf{92.81}  \\

  \bottomrule
\end{tabular}
\end{table}

\subsection{Ablation Study}

\begin{table}[!t]
\centering
  \caption{An ablation study of \shortname{} on the MRNE-2 dataset.}
  \small
  \label{tab:Ablation}
  \begin{tabular}{@{}ccccc@{}}
    \midrule
    Method & Accuracy & Precision & Recall & F1 \\
    \midrule
    \multicolumn{5}{c}{\textit{Without Real Texts}} \\
    Ours & 92.63 & 87.48 & 84.06 & 85.74  \\
    No CCG & 81.54 & 71.61 & 70.94 & 71.27\\
    CLIP Teacher & 72.12 & 61.50 & 56.41 & 58.84 \\
    Random Teacher & 73.36 & 60.89 & 62.03 & 61.46 \\
    No Teacher & 73.54 & 62.12 & 61.25 & 61.68 \\
    Unimodal & 73.42 & 63.78 & 55.31 & 59.25 \\

    \midrule
    \multicolumn{5}{c}{\textit{Without Real Images}} \\
    Ours & 95.42 & 93.74 & 88.91 & 91.26\\
    No CCG & 78.75 & 69.77 & 69.72 & 68.21 \\
    CLIP Teacher & 75.40 & 66.37 & 58.59 & 62.24 \\
    Random Teacher & 75.71 & 64.79 & 64.69 & 64.74 \\
    No Teacher & 76.21 & 62.83 & 66.97 & 64.84 \\
    Unimodal & 74.42 & 58.58 & 60.25 & 59.40 \\

  \bottomrule
\end{tabular}
\end{table}

We create the following ablated versions of \shortname{} from the $K=1$ condition: (1) No CCG. This is equivalent to setting $K=0$ and stopping the synthesis at $V_0$. (2) CLIP teacher. It replaces our teacher network with a CLIP teacher, which selects synthetic views with the highest CLIP similarity with the corresponding real views and use the result as training data. Every other aspect is the same as \shortname. (3) Random Teacher. It selects synthetic views randomly as training data. (4) No Teacher, which uses all synthetic data indiscriminately to train the student. (5) Unimodal, where only the real views are used in training. 

We show the results in Table \ref{tab:Ablation} and make the following observations. 
First, synthetic data provide enormous benefits. Comparing with using only unimodal data for training, \shortname{} boosts performance by approximately 20\% in accuracy and 30\% in F1 across both settings. The results convincingly demonstrate that sometimes training data can make all the difference in model performance. 
Second, the teacher plays a key role. Our teacher network beats the CLIP teacher by 26.90-29.02\% and the random teacher by 24.28-26.52\%. Surprisingly, the CLIP teacher does not perform better than the random teacher. This result highlights the importance of filtering synthetic data based on their mutual information with the label as compared to filtering based on CLIP similarity. 
Third, CCG by itself contributes over 10\% improvements in accuracy and F1. Together with Tab. \ref{tab:main_results}, we observe that increasing $K$ is generally beneficial, despite the diminishing marginal effect sizes. As discussed in \S \ref{sec:ccg-justification}, one possible reason may be that CCG increases the diversity of the synthetic data.

\subsection{Number of Synthetic Views in Student}
The design of the student network exploits multiple synthetic views per real view, in a manner similar to test-time data augmentation. We investigate how the number of synthetic views used during training and test affects the overall accuracy and F1 score. 

\begin{figure}
  \centering
  \includegraphics[width=1.0\linewidth]{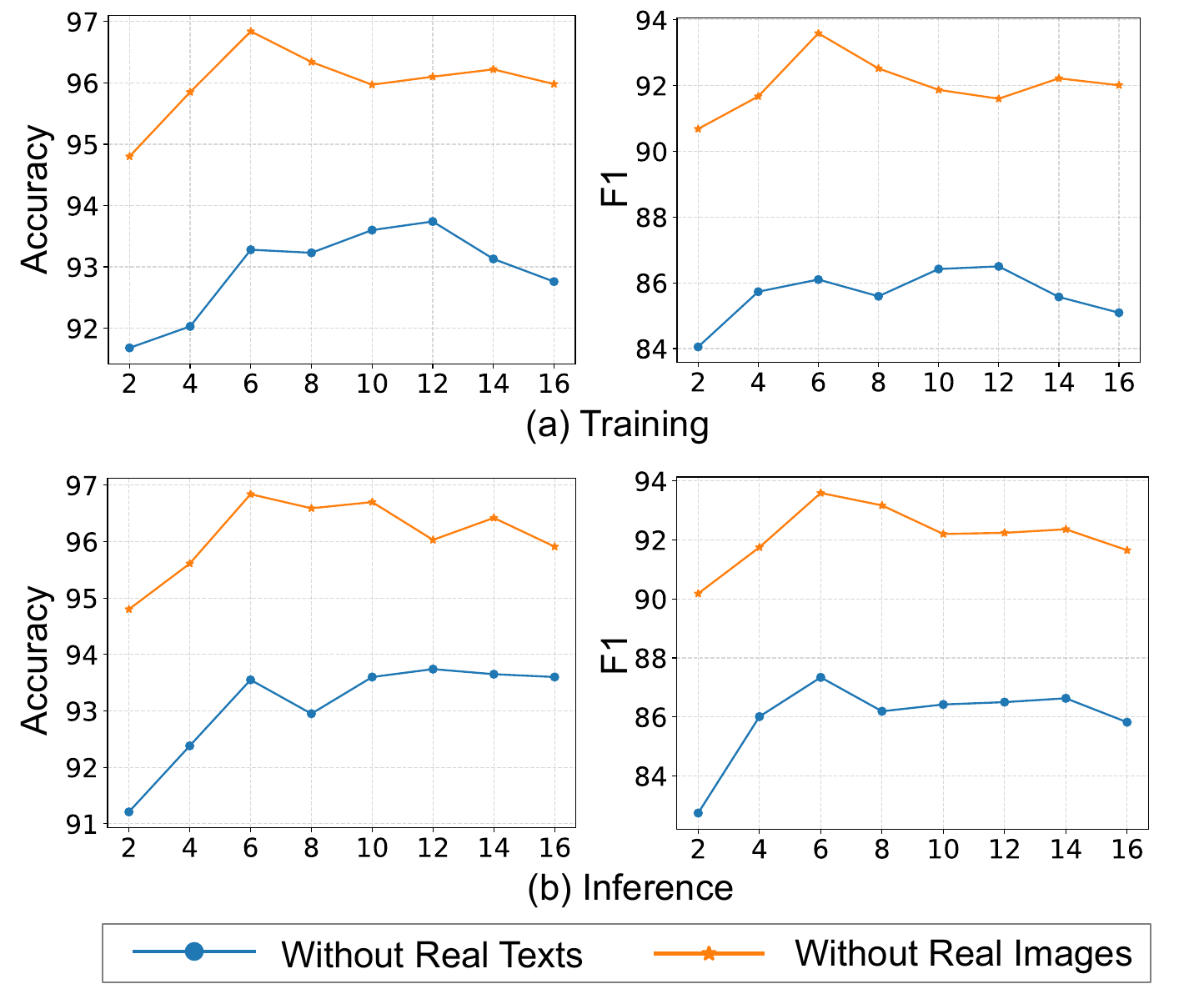}
  \caption{Accuracy and F1 on MNRE-2 as the number of synthetic views used by the student network is varied.}
  \label{IMGscale}
\end{figure}

The top row of Fig. \ref{IMGscale} shows the result when training on a varying number of synthetic views. During inference, we use the exact same number of synthetic views to minimize any distribution shift. We observe a inverse U-shape, where the performance increases to a point (6 synthetic images and 12 synthetic text snippets) and then decreases. 
Performance tends to be low when only 2-4 synthetic views are used, suggesting the potential of synthetic data is not fully realized if insufficient number of views are utilized in training. 

After training completes,  at test time we also vary the number of synthetic views that we create from each test data point. The results are shown on the bottom row of Fig. \ref{IMGscale}. We observe similar trends as the top row. Since each synthetic view is a possible interpretation of the real view, a sufficient number of synthetic views may increase the probably that at least one interpretation is correct and correlates well with the label. 

\subsection{Diversity Estimates}
We aim to verify if CCG indeed improves data diversity, as conjectured in \S \ref{sec:ccg-justification}. However, data diversity is inherently difficult to define because simple statistics may not correspond well to semantic diversity. As an approximate measure, we use the generalized variance of a Gaussian mixture model (GMM) fitted on synthetic data.  First, we extract feature vectors using the original CLIP-ViT-B-32 for synthetic images and BERT-base-uncased for captions. After that, as the data lie in a low-dimensional manifold, we perform dimensionality reduction using principal component analysis. We then fit a 3-component GMM and calculate the generalized variance. We present the variances in Table \ref{tab:main-Diversity}. Each generation step (\ie, $V_0\rightarrow V_1^{\prime}$ and $V_1\rightarrow V_2^{\prime}$) increases variance, suggesting that CCG enhances data diversity. More details and discussions are in the supplementary.


\begin{table}[!t]
\centering
  \caption{Evaluation of the diversity on various synthetic sets.}
  \footnotesize
  \label{tab:main-Diversity}
  \begin{tabular}{@{}ccccc@{}}
    \midrule
    ~ & $V_0$ & $V_1^{\prime}$ & $V_1$ & $V_2^{\prime}$ \\
    \midrule
    Image & \num{6.95e21} & \num{7.37e21} & \num{5.52e21} & \num{7.37e21} \\
    Text & \num{1.85e32} & \num{3.28e32} & \num{3.04e32} & \num{3.71e32} \\
  \bottomrule
\end{tabular}
\end{table}

\subsection{Extension to the Textual Dataset WebNLG}

We investigate if \shortname{} can successfully work with a textual relation extraction dataset with a large number of relations. We adopt theWebNLG dataset \cite{gardent2017creating}, a text-only relation extraction dataset comprising 171 relation types. Like in MNRE-2, we aim to identify relation types of given entity pairs, rather than identifying the entities. 

The results are presented in Table \ref{tab:WebNLG}. Under this setting, our method continues to surpass unimodal training by 5.24\% in F1, highlighting the effectiveness of synthetic multimodal training. In addition, both the teacher model and CCG exhibit performance improvements. However, the performance gains in this setting are not as pronounced as those in multimodal settings (See Table \ref{tab:Ablation}). We hypothesize that the textual data in WebNLG may be less easy to visualize than the multimodal dataset MNRE-2, leading to a smaller advantage of multimodal classifier. For example, WebNLG contains relations such as \texttt{yearOfConstruction} and \texttt{postalCode}, which are difficult to visualize. Nevertheless, \shortname{} still demonstrates higher performance than a pure unimodal approach. Further, the proposed CCG technique and teacher networks both contribute positively. 

\begin{table}[!t]
\centering 
  \caption{Extension to Textual Dataset WebNLG}
  \small
  \label{tab:WebNLG}
  \begin{tabular}{@{}ccccc@{}}
    \midrule
    Method & Accuracy & Precision & Recall & F1 \\
    \midrule
    Ours & \textbf{98.93} & \textbf{94.86} & \textbf{95.64} &  \textbf{95.12} \\
    No CCG & \underline{97.36} & \underline{92.81} & \underline{93.98} & \underline{93.39}\\
    CLIP Teacher & 96.35 & 89.89 & 90.52 & 90.20 \\
    Random Teacher & 96.61 & 90.75 & 90.84 & 90.79 \\
    No Teacher & 96.48 & 90.76 & 91.32 & 91.04 \\
    Unimodal & 96.54 & 89.71 & 90.05 & 89.88 \\
  \bottomrule
\end{tabular}
\end{table}

\section{Related Work}
\label{sec:related_work}

\noindent
\textbf{Multimodal Relation Extraction.}
Recent studies on multimodal relation extraction (MRE) \cite{zheng2021mnre,zheng2021multimodal,chen2022good,wang2022named,yuan2023joint,li2023analyzing,fu2022knowledge,chen2023unified,hu2023multimodal,chen2023chain,cui2023enhancing,li2023dual,zhao2023tsvfn,yu2020improving, zhang2021multi, hu2023prompt, wu2023information} have demonstrated significant progresses. A common approach is to utilize scene graphs \cite{zheng2021multimodal,wu2023information,yuan2023joint}, which provide structured information about the image. However, \cite{li2023analyzing} points out that coarse-grained alignment of scene graphs may not effectively utilize visual guidance. Therefore, to establish fine-grained correspondence between images and texts, \cite{li2023analyzing,li2023dual,huang2023i2srm} align tokens in texts with visual objects. To mitigate vagueness and misleadingness in relations, \cite{chen2023chain} elicits multi-grain textual knowledge from large language models; \cite{wang2022named,hu2023multimodal} retrieve relevant texts and images from Internet knowledge corpus (\eg, Wikipedia). Additionally, RECK \cite{feng2023towards} leverages ConceptNet to extract conceptual relations.

TMR \cite{zheng-etal-2023-rethinking} is the only existing work that makes use of text-to-image generation in MRE. 
In the ablation study of TMR, the synthetic data accounts for about 2.68\% gains in F1, which roughly matches our experiments when no data filtering by the teacher network is performed (Table \ref{tab:Ablation}). The teacher network of \shortname{} leads to significant performance boosts in addition to that. 

In contrast to prior works that train on well-aligned multimodal data, we investigate a new problem, Multimodal Relation Extraction with a Missing Modality, where data from one modality (image or text) are inaccessible during training.

\vspace{0.1in}
\noindent
\textbf{Training with Synthetic Data.}
Building off image generation and image captioning, training on synthetic data has emerged as a promising research direction \cite{Nikolenko-survey-book-2021,Kishore_2021_ICCV}. In computer vision, \cite{Johnson-Roberson2017,Hinterstoisser2018,Tremblay_2018_CVPR_Workshops} explore the use of synthetic data in object detection; \cite{azizi2023synthetic,yebin2023synaug} demonstrate benefits of complementing real images with synthetic images as training data. 
In language tasks, research examines the generation of additional textual training data \cite{gao2023self,honovich2022unnatural,meng2022generating,meng2022tuning,wang2022self,west2021symbolic,ye2022zerogen}. The pioneering work Vokenization \cite{tan2020vokenization} proposes that language tasks can benefit from visualization of textual tokens, but builds the visualizations through retrieval rather than synthesis. The idea has been extended by researchers synthesizing visual data conditioned on the text for applications such as machine translation \cite{long2021generative,gupta2023cliptrans}, story generation \cite{zhu2023visualize}, procedural planning \cite{lu2023multimodal}, dialogue summarization \cite{tang2023learning} and low-resource language understanding \cite{yang2022z,lu2022imagination}. In long sequence understanding tasks, VCoT \cite{rose2023visual} generates both images and texts at intermediate states to bridge semantic gaps between two specified states.

Differing from previous approaches that target unimodal tasks, \cite{du2023training} target multimodal event extraction and generative images from text and vice versa. 
In this work, we extend direct bidirectional generation to iterating through the text-to-image generator and the image-to-text generator multiple times. 

\vspace{0.1in}
\noindent
\textbf{Learning from Noisy Data.}
Noise in training data is a common phenomenon. 
Many works investigate the issue of label noise, where training data samples are associated with incorrect labels. Theoretical analysis \cite{li2020gradient} and empirical observation \cite{arpit2017closer,song2019does} support the notion that DNNs learn generalizable patterns first and gradually overfit to noisy patterns. Therefore, it is common to select data samples with low training loss in subsequent rounds of training \cite{yao2020searching,shen2019learning,chen2019understanding}. To reduce confirmation bias, \cite{malach2017decoupling,han2018co,yu2019does,li2020dividemix} simultaneously train two networks and combine their predictions to identify clean labels. MentorNet \cite{MentorNet-jiang18c} learns a curriculum that selects training data for a student network. In another line of research, \cite{li2021learning} proposes to handle data noise in the form of corrupted and out-of-distribution inputs.

The use of one neural network to select training data for another~\cite{han2018co, MentorNet-jiang18c} bears resemblance to our approach. However, in our problem setting, the labels are assumed to be always correct. The purpose of our teacher network is to prevent gradual loss of label information in the synthetic data rather than identifying incorrect labels.

\section{Conclusions}
\label{sec:conclusions}

This paper introduces a new problem in Multimodal Relation Extraction: only unimodal data (either text or image) is available during training and we need to generate data for the missing modality to train a MRE classifier. 
To tackle this problem, we propose the \longlongname{} approach that comprises two procedures. Chained Cross-modal Generation promotes data diversity by repeatedly applying the text-to-image and image-to-text generators. To mitigate the label information loss, \shortname{} selects synthetic training data that have higher mutual information with the ground-truth labels. Experiments on the MNRE-2 benchmark demonstrate that the MRE classifier trained using the synthetic data generated by \shortname{} outperforms state-of-the-art models trained on real multimodal data on the real multimodal test set. Ablation studies further underscore the effectiveness of each component in \shortname{}. Additionally, \shortname{} exhibits performance gains in traditional text-only relation extraction by converting unimodal training data to multimodal data. We believe \shortname{} opens new avenues to the MRE problem and training on synthetic data in general. 
{
    \small
    \bibliographystyle{ieeenat_fullname}
    \bibliography{main}
}

\clearpage
\setcounter{page}{1}
\maketitlesupplementary

\section{Mathematical Details}
\label{sec:math_details}
\subsection{Proposition 1}
We have two random variables, input $X$ and a discrete label $Y$. $Y$ can take on $M$ values with equal probability. $f$ is an arbitrary neural network  whose output are not necessarily normalized. We would like to show
\begin{equation} 
I(X, Y) \ge \mathbb{E}_{X, Y}\left[\log \frac{\exp f(x, y)}{\sum_{y^{\prime}}[\exp f(x, y^{\prime})]}\right].
\end{equation}

\vspace{0.1in}
\noindent \emph{Proof.} 
\begin{align} 
I(X, Y) & = \mathbb{E}_{X, Y}\left[\log \frac{P(X, Y)}{P(X)P(Y)}\right] \\
& = \mathbb{E}_{X, Y}\left[\log \left( \frac{Q(X, Y)}{P(X)P(Y)} \frac{P(X, Y)}{Q(X, Y)} \right)\right] \\
& = \mathbb{E}_{X, Y}\left[\log \frac{Q(X, Y)}{P(X)P(Y)}\right] \nonumber \\ 
& \quad \quad + \mathbb{E}_{X, Y}\left[\log \frac{P(X, Y)}{Q(X, Y)}\right]  \\
& = \mathbb{E}_{X, Y}\left[\log \frac{Q(X, Y)}{P(X)P(Y)}\right] \nonumber \\ 
& \quad \quad +  KL[P(X, Y) \| Q(X, Y)] \\
& \ge \mathbb{E}_{X, Y}\left[\log \frac{Q(X, Y)}{P(X)P(Y)}\right] \label{eq:14}
\end{align}
Here $Q(X, Y)$ is an arbitrary distribution. To ensure it is a valid distribution, we let 
\begin{equation}
\label{eq:variational-q}
Q(X=x, Y=y)=\frac{P(x)P(y)\exp(f_{\theta}(x, y))}{\mathbb{E}_{y^\prime}[\exp(f_{\theta}(x, y^\prime))]}
\end{equation}
We can easily verify that 
\begin{equation}
\int_{-\infty}^{\infty} Q(X=x, Y=y) \; dx \; dy = 1
\end{equation}
Substituting Eq. \ref{eq:variational-q} back to Eq. \ref{eq:14}, we obtain
\begin{align} 
I(X, Y) & \ge \mathbb{E}_{X, Y}\left[\log \frac{\exp f(x, y)}{\mathbb{E}_{y^{\prime}}[\exp f(x, y^{\prime})]}\right] \\
& = \mathbb{E}_{X, Y}\left[\log \frac{\exp f(x, y)}{\frac{1}{M} \sum_{y^{\prime}}[\exp f(x, y^{\prime})]}\right] \\
& = \mathbb{E}_{X, Y}\left[\log \frac{\exp f(x, y)}{\sum_{y^{\prime}}[\exp f(x, y^{\prime})]}\right] + \log M \\
& \ge \mathbb{E}_{X, Y}\left[\log \frac{\exp f(x, y)}{\sum_{y^{\prime}}[\exp f(x, y^{\prime})]}\right] \rlap{$\qquad \Box$}
\end{align}

\subsection{Proposition 2}
\label{sec:mi-proof}
Suppose we have a sequence of random variables $X_1, X_2, \ldots, X_N$ and they form a Markov chain. In other words, for all integer $M$, $X_i$ and $X_{i+M}$ are independent given $X_{i+m}, \forall 1 \le m \le M-1$. We want to prove that the mutual information $I(X_1; X_N) \le I(X_i, X_j), \forall 1 < i < j \le N$. 

\vspace{0.1in}
\noindent \emph{Proof.} First, consider $X_1, X_i$ and $X_N, i<N$. By the chain rule of mutual information,
\begin{equation}
\begin{split}
    I(X_1 X_i; X_N) & = I(X_1; X_N) + I(X_i;X_N|X_1) \\
    & = I(X_i; X_N) + I(X_1;X_N|X_i) 
\end{split}
\end{equation}
Since $X_1$ and $X_N$ are independent given $X_i$, $I(X_1;X_N|X_i) = 0$. 
\begin{equation}
I(X_1; X_N) = I(X_i; X_N) - I(X_i;X_N|X_1) \le I(X_i; X_N)
\end{equation}
Next, we consider $I(X_N X_j; X_i), i<j<N$. By the same logic as above,
\begin{equation}
I(X_i; X_N) \le I(X_i; X_j), \forall i < j < N. 
\end{equation}
This completes the proof. \qed

\section{Diversity Estimates}


\begin{table*}[!t]
\centering
  \caption{Evaluation of the diversity on various synthetic sets.}
  \small
  \label{tab:Diversity}
  \begin{tabular}{@{}ccccccc@{}}
    \midrule
    ~ & $D_\text{PCA}$ & $N$ & $V_0$ & $V_1^{\prime}$ & $V_1$ & $V_2^{\prime}$ \\
    \midrule
    \multirow{9}{*}{Image} & \multirow{3}{*}{16} & 1 & \num{1.06e13} & \num{1.42e13} ($+33.96$\%) & \num{9.98e12} & \num{1.51e13} ($+51.30$\%) \\
    ~ & ~ & 3 & \num{9.12e12} & \num{1.20e13} ($+31.58$\%) & \num{8.69e12} & \num{1.32e13} ($+51.90$\%) \\
    ~ & ~ & 5 & \num{7.11e12} & \num{1.15e13}  ($+61.74$\%) & \num{6.71e12} & \num{1.17e13}  ($+74.37$\%) \\
    \cmidrule{2-7}
    ~ & \multirow{3}{*}{32} & 1 & \num{9.12e21} & \num{9.36e21} ($+2.63$\%) & \num{6.47e21} & \num{9.55e21} ($+47.60$\%) \\
    ~ & ~ & 3 & \num{6.95e21} & \num{7.37e21} ($+6.04$\%) & \num{5.52e21} & \num{7.37e21} ($+33.51$\%) \\
    ~ & ~ & 5 & \num{5.95e21} & \num{6.74e21}  ($+13.28$\%) & \num{4.57e21} & \num{7.12e21}  ($+55.80$\%) \\
    \cmidrule{2-7}
    ~ & \multirow{3}{*}{64} & 1 & \num{3.08e34} & \num{1.47e34}  ($-52.27$\%) & \num{1.23e34} & \num{1.41e34}  ($+14.63$\%) \\
    ~ & ~ & 3 & \num{2.32e34} & \num{1.24e34}  ($-46.55$\%) & \num{1.01e34} & \num{1.18e34}  ($+16.83$\%) \\
    ~ & ~ & 5 & \num{2.17e34} & \num{1.03e34}  ($-52.53$\%) & \num{8.68e33} & \num{9.58e33}  ($+10.37$\%) \\
    \midrule
    \multirow{9}{*}{Text} & \multirow{3}{*}{16} & 1 & \num{6.43e20} & \num{8.75e20}  ($+36.08$\%) & \num{8.38e20} & \num{1.00e21}  ($+19.33$\%) \\
    ~ & ~ & 3 & \num{5.75e20} & \num{7.77e20}  ($+35.13$\%) & \num{7.45e20} & \num{9.06e20}  ($+21.61$\%) \\
    ~ & ~ & 5 & \num{5.81e20} & \num{7.53e20}  ($+29.60$\%) & \num{7.39e20} & \num{8.72e20}  ($+18.00$\%) \\
    \cmidrule{2-7}
    ~ & \multirow{3}{*}{32} & 1 & \num{2.17e32} & \num{3.86e32}  ($+77.88$\%) & \num{3.57e32} & \num{4.22e32}  ($+18.21$\%) \\
    ~ & ~ & 3 & \num{1.85e32} & \num{3.28e32}  ($+77.30$\%) & \num{3.04e32} & \num{3.71e32}  ($+22.04$\%) \\
    ~ & ~ & 5 & \num{1.74e32} & \num{2.86e32}  ($+64.37$\%) & \num{2.91e32} & \num{3.43e32}  ($+17.87$\%) \\
    \cmidrule{2-7}
    ~ & \multirow{3}{*}{64} & 1 & \num{2.45e43} & \num{1.06e44}  ($+332.65$\%) & \num{1.02e44} & \num{7.96e43}  ($-21.96$\%) \\
    ~ & ~ & 3 & \num{2.04e43} & \num{8.68e43}  ($+325.49$\%) & \num{8.57e43} & \num{6.49e43}  ($-24.27$\%) \\
    ~ & ~ & 5 & \num{1.85e43} & \num{7.84e43}  ($+323.78$\%) & \num{7.43e43} & \num{5.94e43}  ($-20.05$\%) \\
  \bottomrule
\end{tabular}
\end{table*}

Data diversity is inherently difficult to define because simple statistics may not correspond well to semantic diversity. As an approximate measure, we use the generalized variance of a Gaussian mixture model (GMM) fitted on synthetic data. First, we extract a feature vector from each synthetic sample using a pretrained neural network. We use CLIP-ViT-B-32 as the image encoder and BERT-base-uncased as the text encoder. As the data lie in a low-dimensional manifold, we apply principal component analysis (PCA) to reduce the feature dimension to $D_{\text{PCA}}$. Second, we fit a GMM, which contains $N$ Gaussian distributions with mean $\bm u_i$ and diagonal covariance matrix $\Sigma_i$.

To calculate the variance of the GMM, we apply the law of total variance. Let random variable $C$ denote the component that a data point $X$ belongs to, we have
\begin{equation}
\begin{split}
\text{Var}(X)& =\mathbb{E}[\text{Var}(X|C)]+\text{Var}(\mathbb{E}[X|C]) \\
& =\sum_i^N p_i \boldsymbol{\Sigma}_i+\sum_i^N p_i (\boldsymbol{u}_i-\bar{\boldsymbol{u}})(\boldsymbol{u}_i-\bar{\boldsymbol{u}})^{\top}, \\
\end{split}
\end{equation}
since $\text{Var}(X|C=i)=\Sigma_i$ and $\mathbb{E}_X[X|C=i]=\bm{u_i}$. $\bar{\boldsymbol{u}}=\sum_i^N p_i \boldsymbol{u}_i$.
We use generalized variance, which is the determinant of the covariance matrix. 
\begin{equation}
\sigma^2=\text{det}\left(\sum_i^N p_i \boldsymbol{\Sigma}_i+\sum_i^N p_i (\boldsymbol{u}_i-\bar{\boldsymbol{u}})(\boldsymbol{u}_i-\bar{\boldsymbol{u}})^{\top}\right)
\end{equation}


In Table \ref{tab:Diversity}, we present the variance results in two CCG steps using different numbers of GMM components and feature dimensions. Across different choices of GMM components, we consistently observe that variance increases with each generation step when $D_\text{PCA}$ is 16 or 32. However, when the $D_\text{PCA} = 64$, this trend does not always hold true. We argue that this shows CCG increases variance on the most important principal components of the feature space. Variations in the less important principal directions are likely less recognizable by humans and could be considered as noise. Examples of such variations are changes in the blurred image regions or textures of grass and tree leaves. Though humans tend to ignore such details, they contribute to the variance nonetheless. However, such details probably do not contribute significantly toward representation learning of the student network. 



\section{Implementation Details}

\subsection{Experiments on MNRE}
During training of the student and teacher, the optimizer and learning rate scheduler is AdamW and Cosine Decay. The student adopts a learning rate of $10^{-5}$, a batch size of 16, and a weight decay of $10^{-2}$, while the teacher employs a batch size of 256 and a weight decay of $10^{-2}$. The learning rate of the teacher is set to $3\times10^{-7}$ and $5\times10^{-7}$ when training on synthetic texts and images, respectively. 

Data augmentation is applied to real data during student training to mitigate overfitting. For the real texts, we randomly select 30\% tokens excluding the entity pair. Then, we mask them with the probability of 0.8, or replace them with a random token in the vocabulary with the probability of 0.2. For the real images, we apply the RandAugment \cite{cubuk2020randaugment}, where  the number of augmentation transformations and the magnitude for transformations is 2 and 7, respectively.

\subsection{Experiments on WebNLG}

We set the number of Chained Cross-modal Generation (CCG) iterations, denoted as $K$, to 1. For each real text in $U_0$, we generate 10 synthetic images for $V_0^\prime$. After the teacher network selects 60\% images (i.e., 6 images), the first CCG round generates 24 synthetic views, providing 4 new images for each input. At last, we choose 4 synthetic images for each real text during student training. 

In the student model, we adopt a learning rate of $5\times10^{-5}$, a batch size of 16, and a weight decay of $10^{-2}$. For the teacher model, we use a learning rate of $7\times10^{-6}$, a batch size of 128, and a weight decay of $10^{-2}$. To address potential overfitting issues, we implement the same text augmentation as used in MNRE.

\section{Dataset Statistics}
The statistics of the multimodal dataset MNRE compared with the textual dataset WebNLG are listed in Tab. \ref{tab:Statistics}. Sentences in WebNLG contain 68\% more words than those in MNRE. This difference in text length suggests sentences in WebNLG are not easily visualized, potentially reducing the advantage of the multimodal classifier.

\begin{table}[!t]
\centering 
  \caption{Statistics of MNRE and WebNLG}
  \small
  \label{tab:Statistics}
  \begin{tabular}{@{}lcc@{}}
    \midrule
    Statistics & MNRE & WebNLG \\
    \midrule
    \# Word & 258k & 290k \\
    \# Sentence & 9,201 & 6,222 \\
    \# Words Per Sentence & 28 & 47 \\
    \# Instance & 15,485 & 14,485 \\
    \# Entity & 30,970 & 28,970 \\
    \# Relation & 23 & 171 \\
    \# Image & 9,201 & - \\
    
  \bottomrule
\end{tabular}
\end{table}

\section{Baselines in Main Results}

We compare \shortname{} against 19 baselines for the multimodal relation extraction. It is noted that all of them are trained on real multimodal data. Here we give more details.

The methods in the first category leverage a large number of image-text pairs. TMR \cite{zheng-etal-2023-rethinking} and \cite{hu2023prompt} additionally leverage 400k and 988k unlabeled image-caption pairs, repectively. In addition, TMR \cite{zheng-etal-2023-rethinking} utilizes text-to-image generation, employing real text and generated images for multi-grained representation learning. In the second category, image and text retrieval is employed. For example, MoRe \cite{wang2022named} and \cite{hu2023multimodal} enhance their datasets by retrieving additional images and texts from Wikipedia and Google, respectively. GPT4-XMLR \cite{chen2023chain} retrieves textual information from large language models to train a compact student model. 

Other works can be further categorized into two groups. In the first group, MRE-ISE \cite{wu2023information} and MEGA \cite{zheng2021multimodal} utilize pretrained scene graph extraction models, employing graph neural networks to obtain multimodal graph representations for structural alignment. In the second group, HVPNet \cite{chen2022good}, DGF-PT \cite{li2023dual}, I$^2$SRM \cite{huang2023i2srm}, Iformer \cite{li2023analyzing}, and RECK \cite{feng2023towards} integrate pretrained object detectors to achieve fine-grained alignment between textual tokens and visual objects. 


\section{Comparison Against Test-time Data Augmentation}

In this section, we compare our proposed method with a similar approach, test-time data augmentation (TDA), which also leverages multiple different images as an ensemble during inference. Specifically, the model takes one image-text pair at a time and is trained on real data in MNRE. During inference, we use RandAugment \cite{cubuk2020randaugment} on the real image and randomly obtain $N$ augmented images. We collect these synthetic images and the real images into an image pool. Subsequently, we feed the real text and one image from the pool to the model at a time, yielding $N+1$ probability vectors, which are averaged for the final prediction. The label with the highest probability is chosen as the ensemble prediction. We set $N$ to 6, the same as the student network in \shortname{}. The results are presented in Table \ref{tab:test}. We observe that TDA does not yield performance gain and even marginally decreases performance. 


\begin{table}[!t]
\centering 
  \caption{Experimental results of test-time data augmentation. There are two hyperparameters $\lambda_1$ and $\lambda_2$ in the RandAugment. $\lambda_1$ represents the number of augmentation transformations to apply sequentially, while $\lambda_2$ corresponds to the magnitude for all the transformations in the sequence.}
  \small
  \label{tab:test}
  \begin{tabular}{@{}ccccc@{}}
    \midrule
    Method & Accuracy & Precision & Recall & F1 \\
    \midrule
    \shortname{} & \textbf{96.65} & \textbf{93.92} & \textbf{91.72} & \textbf{92.81} \\
    No TDA & 76.77 & 65.63 & 65.94 & 65.78  \\
    $\lambda_1$=1, $\lambda_2$=3 & 76.64 & 65.67 & 65.47 & 65.57 \\
    $\lambda_1$=2, $\lambda_2$=5 & 76.58 & 65.87 & 64.84 & 65.35\\
    $\lambda_1$=3, $\lambda_2$=7 & 76.64 & 65.77 & 65.16 & 65.46 \\
    $\lambda_1$=4, $\lambda_2$=9 & 76.58 & 65.97 & 64.53 & 65.24\\
  \bottomrule
\end{tabular}
\end{table}

\section{Examples of cross-modality generation}

\begin{figure*}[t]
  \centering
  \includegraphics[width=1.0\linewidth]{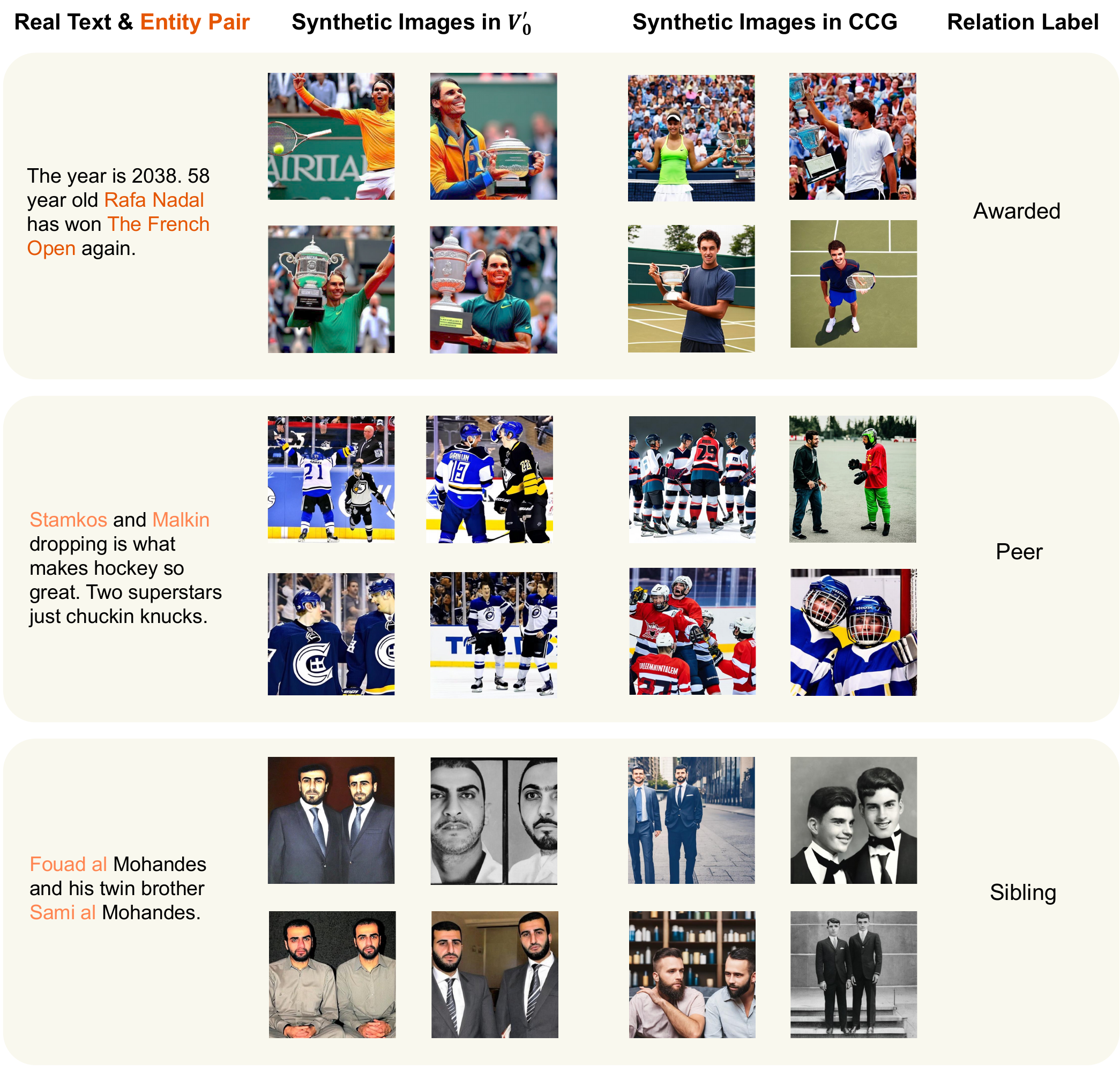}
  \caption{Examples of generated images when the real image are missing. We show the images in $V_0^\prime$ which is directly generated from the real text by text-to-image generation. In addition, we give some synthetic samples in the Chained Cross-modality Generation (CCG).}
  \label{IMGExamples_T2I}
\end{figure*}

\begin{figure*}[t]
  \centering
  \includegraphics[width=1.0\linewidth]{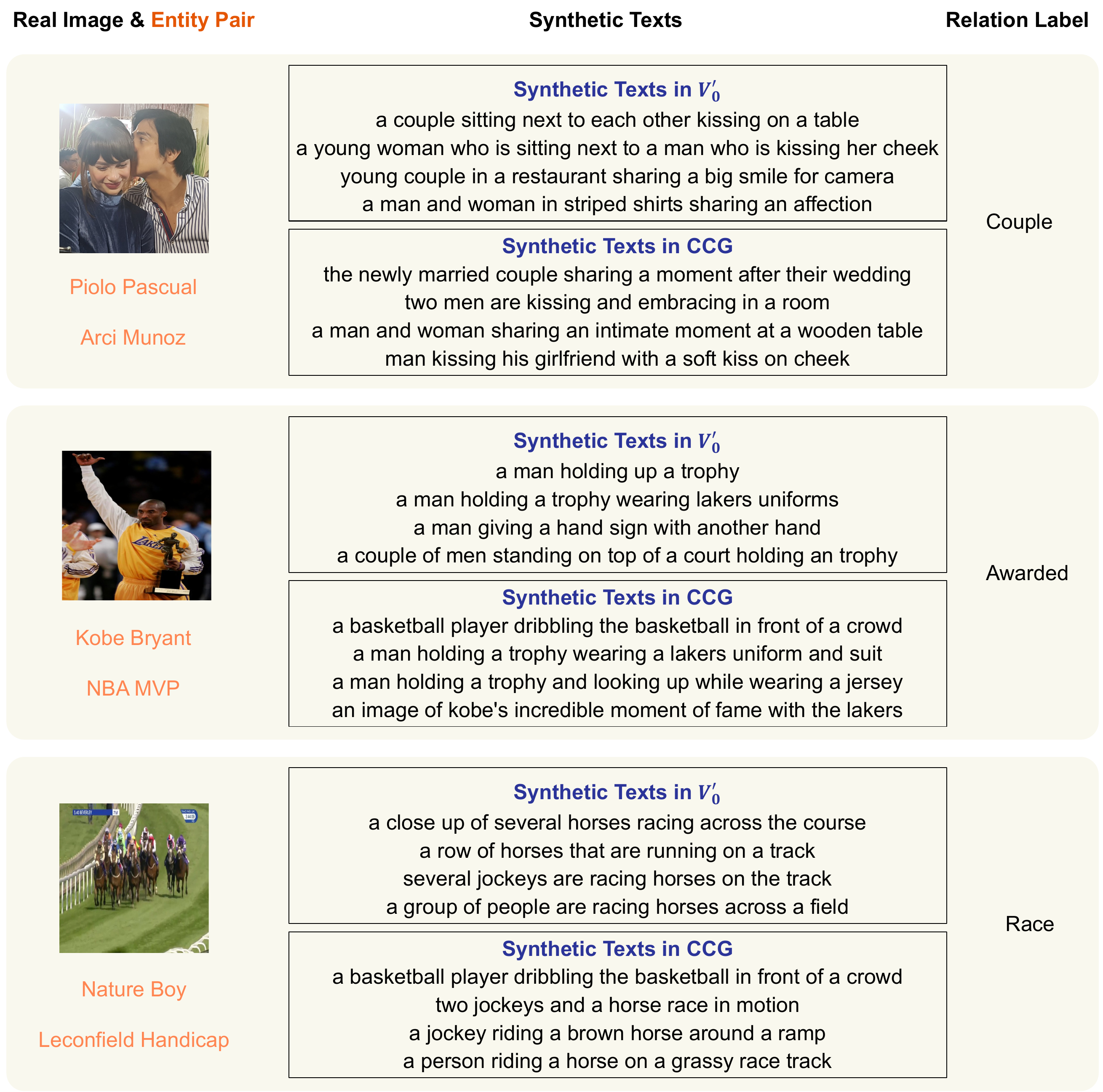}
  \caption{Examples of generated captions when the real text are missing. We present the captions in $V_0^\prime$ which is directly generated from the real image by image-to-text generation. Besides, we show some synthetic caption in the Chained Cross-modality Generation (CCG).}
  \label{IMGExamples_I2T}
\end{figure*}

We give more synthetic examples in Fig. \ref{IMGExamples_T2I} and \ref{IMGExamples_I2T}. We can observe that the synthetic data generated in CCG is more diverse compared to those directly generated from real data. 

\end{document}